\documentclass[10pt,twocolumn,letterpaper]{article}

\usepackage{iccv}
\usepackage{times}
\usepackage{epsfig}
\usepackage{graphicx}
\usepackage{amsmath}
\usepackage{amssymb}
\usepackage{multirow}
\usepackage{booktabs}
\usepackage{url}

\usepackage[pagebackref=true,breaklinks=true,letterpaper=true,colorlinks,bookmarks=false]{hyperref}

\iccvfinalcopy 

\def\iccvPaperID{7193} 

\ificcvfinal\fi

\begin{document}

\title{\vspace{-25pt}Joint-Relation Transformer for Multi-Person Motion Prediction}
\author{Qingyao Xu\textsuperscript{1\footnotemark[1]}, \quad Weibo Mao\textsuperscript{1\footnotemark[1]}, \quad Jingze Gong\textsuperscript{1}, \quad Chenxin Xu\textsuperscript{1},\\ Siheng Chen\textsuperscript{1,2\footnotemark[2]}, \quad Weidi Xie\textsuperscript{1,2}, \quad Ya Zhang\textsuperscript{1,2}, \quad Yanfeng Wang\textsuperscript{1,2\footnotemark[2]} \\
$^{1}$Shanghai Jiao Tong University \quad $^{2}$Shanghai AI Laboratory\\
{\tt\small \{xuqingyao, kirino.mao, xcxwakaka, sihengc, weidi, ya\_zhang, wangyanfeng\}@sjtu.edu.cn }\\
{\tt\small autubot@163.com}
}
\maketitle
\renewcommand{\thefootnote}{\fnsymbol{footnote}}
\footnotetext[1]{Equal contribution. \qquad \footnotemark[2]Corresponding author.}
\ificcvfinal\thispagestyle{empty}\fi

\begin{abstract}
  
\end{abstract}

Multi-person motion prediction is a challenging problem due to the dependency of motion on both individual past movements and interactions with other people. Transformer-based methods have shown promising results on this task, but they miss the explicit relation representation between joints, such as skeleton structure and pairwise distance, which is crucial for accurate interaction modeling. In this paper, we propose the Joint-Relation Transformer, which utilizes relation information to enhance interaction modeling and improve future motion prediction. Our relation information contains the relative distance and the intra-/inter-person physical constraints. To fuse relation and joint information, we design a novel joint-relation fusion layer with relation-aware attention to update both features. Additionally, we supervise the relation information by forecasting future distance. 
Experiments show that our method achieves a 13.4\% improvement of 900ms VIM on 3DPW-SoMoF/RC and 17.8\%/12.0\% improvement of 3s MPJPE on CMU-Mpcap/MuPoTS-3D dataset. 
{\em Code is available at \href{https://github.com/MediaBrain-SJTU/JRTransformer}{https://github.com/MediaBrain-SJTU/JRTransformer}.}

\section{Introduction}

Multi-person motion prediction aims to predict the future positions of skeleton joints for multiple individuals based on their historical movements. Compared to traditional single-person motion prediction~\cite{fragkiadaki2015recurrent,chiu2019action, butepage2017deep, li2018convolutional, wei2019motion, wei2020his, li2020dynamic}, multi-person motion prediction is more practical as people are mostly associated with a group and interacting with each other.  It is also more challenging because sophisticated interactions across different individuals need to be considered. The related methods are playing significant roles in a wide range of practical applications, including autonomous driving~\cite{gong2011multi, tang2023collaborative, fang2023tbp,xu2022groupnet}, surveillance systems~\cite{gaur2011string, vu2020anomaly, xu2022remember} and healthcare monitoring~\cite{wagner2018targeted}. They also pave a path to better human-robot interaction~\cite{chen2019crowd, gao2021human}.

Previous works on multi-person motion prediction generally adopt two types of architectures, including graph neural networks (GNNs) and Transformer.  The former one, such as TRiPOD~\cite{adeli2021tripod}, models the multi-person interaction via a graph structure, however, it suffers from the inherent oversoomthing problem in GNNs, thus can only afford shallow models  with limited learning capacities; while the latter one, such as MRT~\cite{wang2021multi} and SoMoFormer~\cite{vendrow2022somoformer}, treat temporal sequence or skeleton joints as a sequence input, 
and learn to establish the relations between them via self-attention mechanism.
Compared with GNN-based methods, Transformer has shown strong learning ability, thus becoming a {\em default} backbone for multi-person motion prediction.


\begin{figure}[t]
\vspace{-15pt}
\begin{center}
   \includegraphics[width=1.0\linewidth]{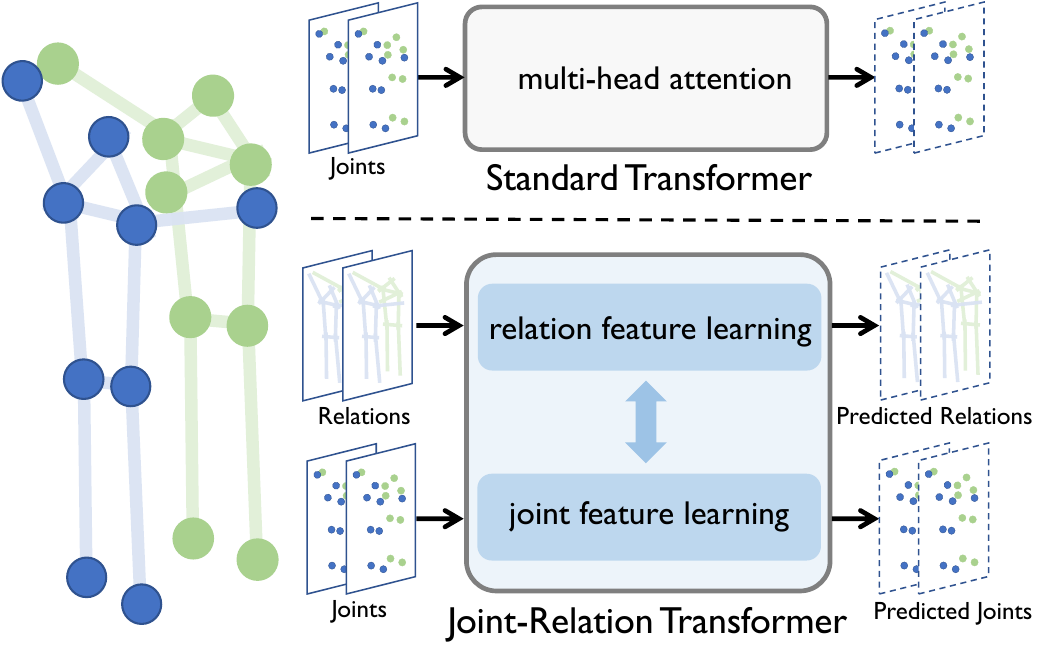}
\end{center}
\vspace{-15pt}
   \caption{System comparison between the standard Transformer and the proposed Joint-Relation Transformer.}
\vspace{-18pt}
\label{fig:system_comparison}
\end{figure}

While encouraging results are shown in the Transformer-based methods, those previous works only implicitly learn the inter-joint relation through the attention mechanism and lack the explicit awareness of skeleton structure. To address this issue, we propose Joint-Relation Transformer, a two-stream Transformer architecture for multi-person motion prediction. Instead of only updating the features for body joints, Joint-Relation Transformer uses two streams to achieve feature learning for both joints and relations. In specific, one stream encodes the sequence of skeleton joints, containing the historical movement information of the joints in world coordinates; and the other one explicitly takes the joints-to-joints relation as input, including the relative distance between two joints and physical constraints, 
such as inter-body constraints and intra-body skeleton connections.

To effectively fuse and update feature of the joints and relation branches, 
we design a novel relation-aware attention to update joints' features with the incorporation of relation features. During the update procedure, attention scores between skeleton joints are calculated from two sources: the similarity score between two joints' features and
the additional relation score based on the relation feature between these two joints. This approach enhances the model's ability to distinguish between similar joint features belonging to distinct persons, leading to more accurate joint updates. Moreover, this design allows for increased granularity in attention allocation, enabling the concentration on the most pertinent joint features for each person, consequently enhancing prediction performance.

To train our proposed Joint-Relation Transformer, 
in addition to inferring joint positions, 
we also supervise the model by predicting future inter-joint distances, 
which contains the future relationship between two joints. 
This relative distance supervision is translation and rotation invariant of the input sequence, adhering to invariance properties of multi-person interactions. 
As to evaluate the effectiveness of our method, 
we conduct experiments on four multi-person motion prediction datasets: 3DPW-SoMoF~\cite{adeli2021tripod}, 3DPW-SoMoF/RC, CMU-Mocap~\cite{cmu-mocap}, 
and MuPoTS-3D~\cite{singleshotmultiperson2018}. The quantitative results show we outperform the previous methods and achieve state-of-the-art performance on most datasets. 
The qualitative results verify the reasonable attention allocation and vivid predictions.

To summarise, in this paper, we make the following contributions: 
(i) We propose the Joint-Relation Transformer for multi-person motion prediction. We innovatively introduce the relation information, which explicitly builds the relationship between joints of the inter-/intra-body;
(ii) We design a relation-aware attention module to update the joint information with the corporation of explicit relation information, increasing granularity in attention allocation and enhancing prediction performance;
(iii) We further supervise the relation information between two joints with the future relative distance to better capture the interaction information hidden in the distance variation;
(iv) We perform our experiments on several common datasets and our proposed method outperforms most state-of-the-art methods. We also conduct thorough ablation study to show the importance of the proposed relation information and relation-aware attention in the task of multi-person motion estimation.






\section{Related Work}
\noindent \textbf{Single person motion prediction.} 
To forecast motions, traditional methods typically employ Hidden Markov Models~\cite{lehrmann2014efficient}, Gaussian-process~\cite{wang2005gaussian} and other methods with hand-crafted features~\cite{wu2014leveraging}, etc.
With the development of deep learning, many RNN-based networks~\cite{fragkiadaki2015recurrent, jain2016structural, ghosh2017learning, pavllo2018quaternet, gopalakrishnan2019neural, chiu2019action, liu2019towards, li2021skeleton} have been developed and achieve great success. For instance, ERD~\cite{fragkiadaki2015recurrent} uses an RNN architecture with nonlinear encoder and decoder networks to predict human motion, while TP-RNN~\cite{chiu2019action} captures the latent hierarchical structure of human poses at different time scales via RNN for a better prediction. Despite their success, RNN-based networks suffer from error accumulation problems inherent to their architecture. Some works~\cite{butepage2017deep, li2018convolutional, wei2019motion, wei2020his, li2020dynamic, li2021symbiotic, li2021multiscale, xu2023eqmotion} utilize feed-forward network to alleviate this problem. For example, LTD~\cite{wei2019motion} adopts DCT to encode temporal information and adopts GCN to learn graph connectivity automatically. HRI~\cite{wei2020his}
introduces an attention-based feed-forward network to extract motion attention between sub-sequences. DMGNN~\cite{li2020dynamic} introduces a multiscale graph to comprehensively model the internal relations of human motion. EqMotion\cite{xu2023eqmotion} proposes an efficient equivariant motion prediction model to maintain motion equivariance and interaction invariance. However, these methods primarily concern situations involving only one person, lacking the consideration of interaction in real-world scenarios.

\vspace{3pt}
\noindent \textbf{Multi-person human motion prediction.}
Recent studies have emphasized multi-person forecasting problems. Due to the existence of interactive behavior, people's motions are likely to be affected by others. To handle this problem, JSC~\cite{adeli2020socially} models global motion and local body joint movements separately with a shared GRU encoder to incorporate scene and social contexts. TRiPOD~\cite{adeli2021tripod} adopts GAT which considers the person and objects as graph nodes to capture the interaction between them. Nowadays many works~\cite{wang2021multi, guo2022multi, vendrow2022somoformer} start to explore the application of Transformers in this task because of its powerful learning ability. MRT~\cite{wang2021multi} pays attention to the change and correlation of human pose in time and uses a global encoder to capture the overall interaction between humans of each frame. SoMoFormer~\cite{vendrow2022somoformer} carefully models the interaction between all joints, but simply uses the Transformer's attention mechanism and lacks a detailed design for relation modeling. DuMMF\cite{xu2023stochastic} focuses on stochastic predicting and proposes a dual-level framework to tackle local individual motion and global social interactions. We notice that although achieving a superior result, Transformer-based methods only use the attention mechanism to implicitly infer the joints' interaction, missing the possible explicit relation information between joints. To overcome this problem, we introduce the relation information into the Transformer architecture and propose Joint-Relation Transformer that can tackle and utilize both the joint information and relation information.

\begin{figure*}[!t]
\begin{center}
\includegraphics[width=1.0\linewidth]{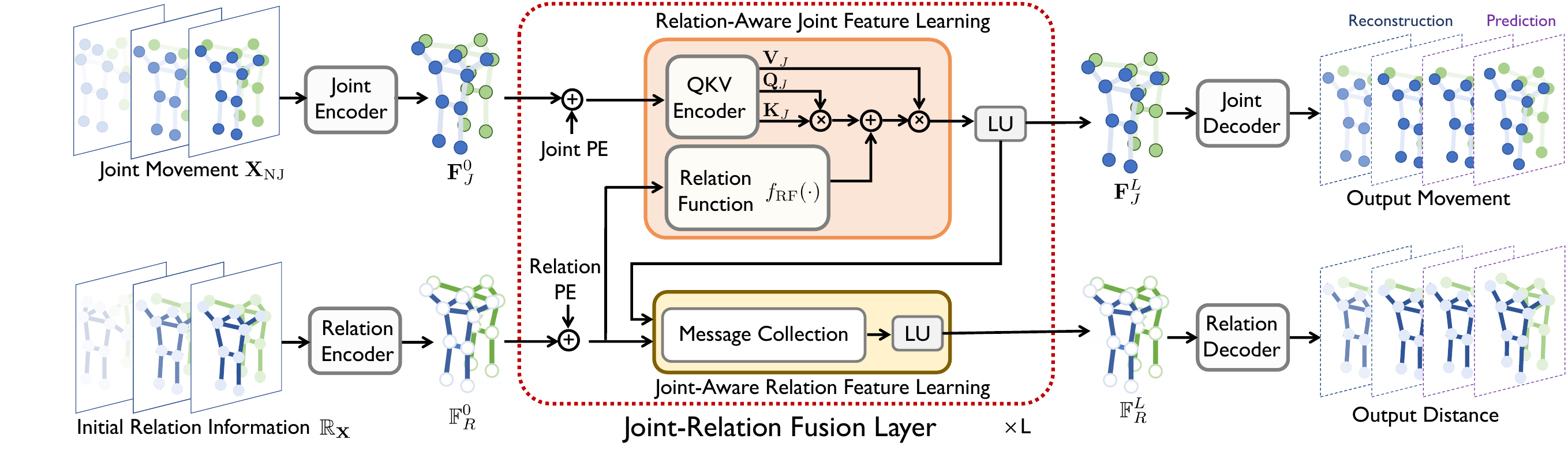}
\end{center}
\vspace{-10pt}
   \caption{Architecture of proposed Joint-Relation Transformer. The network contains three stages: i) encoder modules to extract features, ii) a fusion module to fuse the joint and relation features, and iii) decoder modules to output final results. LU: local update module.}
\vspace{-15pt}
\label{fig:framework}
\end{figure*}

\vspace{3pt}
\noindent \textbf{Graph transformer.}
Transformer~\cite{vaswani2017attention} has made great success in many fields and shown great potential in modeling graph data~\cite{hu2021ogb, min2022masked, chanussot2021open}. Many methods have been proposed to incorporate graph structure into Transformer,  which can be divided into three categories. The first one is to adopt GNN as an auxiliary module where the attention block and the GNN block are always separate~\cite{rong2020self, zhang2020graph, mialon2021graphit, lin2021mesh, wu2021representing}. The second one uses the improved positional embedding derived from graph structure to have a better perception of graph structure information~\cite{dwivedi2020generalization, hussain2021edge, kreuzer2021rethinking, ying2021transformers}. The third one focus on the attention matrix design and uses graph structure to get a more representative matrix~\cite{khoo2020interpretable, zhao2021gophormer, diao2022relational}. Our method is more related to the third kind where we design a special attention mechanism to calculate relation-aware attention score. 



\section{Problem Formulation}



Multi-person motion prediction aims to predict the future positions of joints for multiple individuals based on their historical movements. Mathematically, given a scene with $N$ persons, each has $J$ skeleton joints, let $\mathbf{X}_{\rm NJ} =[X_1, X_2, \cdots, X_{NJ} ]\in \mathbb{R}^{NJ \times (T_h \times 3)}$ be the observed sequence in the scene over $T_h$ history timestamps, 
where $X_{nj} \in \mathbb{R}^{T_h \times 3}$ refers to the observed joint sequence of the $j$-th joint on $n$-th person in 3D world coordinate. 
Similarly, we define the future motions as ${\mathbf{Y}}_{\rm NJ}=[{Y}_1, {Y}_2, \cdots, {Y}_{NJ} ]\in \mathbb{R}^{NJ \times (T_f \times 3)}$, and ${Y}_{nj} \in \mathbb{R}^{T_f \times 3}$ represents the future joint sequence for the $j$-th joint of $n$-th person. 
The goal is to train a computational model $\mathcal{F}(\cdot)$ that infers the future motions of multiple individuals by the observed motion $\widehat{\mathbf{Y}}_{\rm NJ} = \mathcal{F}(\mathbf{X}_{\rm NJ})$ to approximate the ground-truth future motion ${\mathbf{Y}}_{\rm NJ}$. Unless otherwise specified, we will use $\mathbf{X}/\mathbf{Y}$ for ground-truth matrices, $\mathbb{D}_\mathbf{X}/\mathbb{D}_\mathbf{Y}$ for tensors in past and future, and $\widehat{\mathbf{X}}/\widehat{\mathbf{Y}}$ for generated matrices.

This task is challenging for several reasons: 
i) the movement of a person is bounded by physical constraints, {\em e.g.} two joints with a bone connection should maintain the constant relative distance; 
and ii) the behavior of the joints is influenced by both the internal structure of the person's joints, as well as inter-personal joints. To alleviate this dilemma, we introduce the relation information and fuse the information with the proposed Joint-Relation Transformer.

\section{Architecture}
In this section, we present a two-stream architecture for multi-person motion prediction, 
that simultaneously infers the position of joints and their relations/distance,
as shown in Fig.~\ref{fig:framework}.
Mathematically, let $\mathbf{X}_{\rm NJ}$ be the observed joint information,
the \textbf{inference procedure} can be formulated as: 
\begin{equation}
 \widehat{\mathbf{X}}_{\text{NJ}}, \widehat{\mathbf{Y}}_\text{NJ}, \widehat{\mathbb{D}}_\mathbf{X}, \widehat{\mathbb{D}}_\mathbf{Y}  = 
 \Phi_{\text{decode}}(\cdot) \circ \Phi_{\text{fuse}}(\cdot) \circ \Phi_{\text{encode}}(\mathbf{X}_\text{NJ}), 
 \nonumber
\end{equation}
in particular, the historical sequences $\mathbf{X}_\text{NJ}$ are firstly processed with an encoder module $\Phi_{\text{encode}}(\cdot)$~(Sec.~\ref{sec:input}) to obtain the initial joints and relation representation.
Then the output representation is passed into a fusion module $\Phi_{\text{fuse}}(\cdot)$, 
to fuse and update the joints and relation representation~(Sec.~\ref{sec:fusion_layer}).
Lastly, the fused features are processed by the decoder module $\Phi_\text{decode}(\cdot)$, outputting the joints' positions $\widehat{\mathbf{X}}_{\text{NJ}}, \widehat{\mathbf{Y}}_\text{NJ}$ and distance between joints $\widehat{\mathbb{D}}_\mathbf{X}, \widehat{\mathbb{D}}_\mathbf{Y}$~(Sec.~\ref{sec:decoder}).
At training time, the model is optimized to infer motions for both historical and future timestamps, potentially alleviating the catastrophic forgetting issue; 
while at inference time, we only take the predictions for future timestamps~(as detailed in Sec.~\ref{sec:training_objective}).




\subsection{Encoder Module}
\label{sec:input}
In the literature, Transformer-based motion prediction models have primarily considered using the 3D world coordinates of joints as input, like MRT~\cite{wang2021multi} and SoMoFormer~\cite{vendrow2022somoformer},
that requires the model to implicitly learn the complex dependencies between joints.
Here, we propose to enrich the motion representation by augmenting it with 
temporal differentiation and \textbf{explicit joint relations}. 
In particular, we independently process the historical motion sequences by a joint encoder which considers temporal differentiation, and a relation encoder, which considers explicit joint relations, as detailed in the following.


\vspace{5pt}
\noindent \textbf{Joint Encoder.}
Instead of only using the joint position information in world coordinates as input, here we augment the joint information by taking the temporal differentiation between current and one step before: $\Delta X_{nj}^t=X_{nj}^{t}-X_{nj}^{t-1}$, {\em i.e.}, the velocity information and concatenating them. 

Till here, each joint is effectively encoded by a vector of $T_h \times 6$ dimensions. 
We then employ a 2-layer MLP encoder to project each joint feature into a higher dimension. We denote the output of the joint encoder as $\mathbf{F}_J^0\in\mathbb{R}^{NJ\times D}$ with the feature dimension $D$.

 \begin{figure}[t]
\begin{center}
\vspace{-10pt}
\includegraphics[width=1\linewidth]{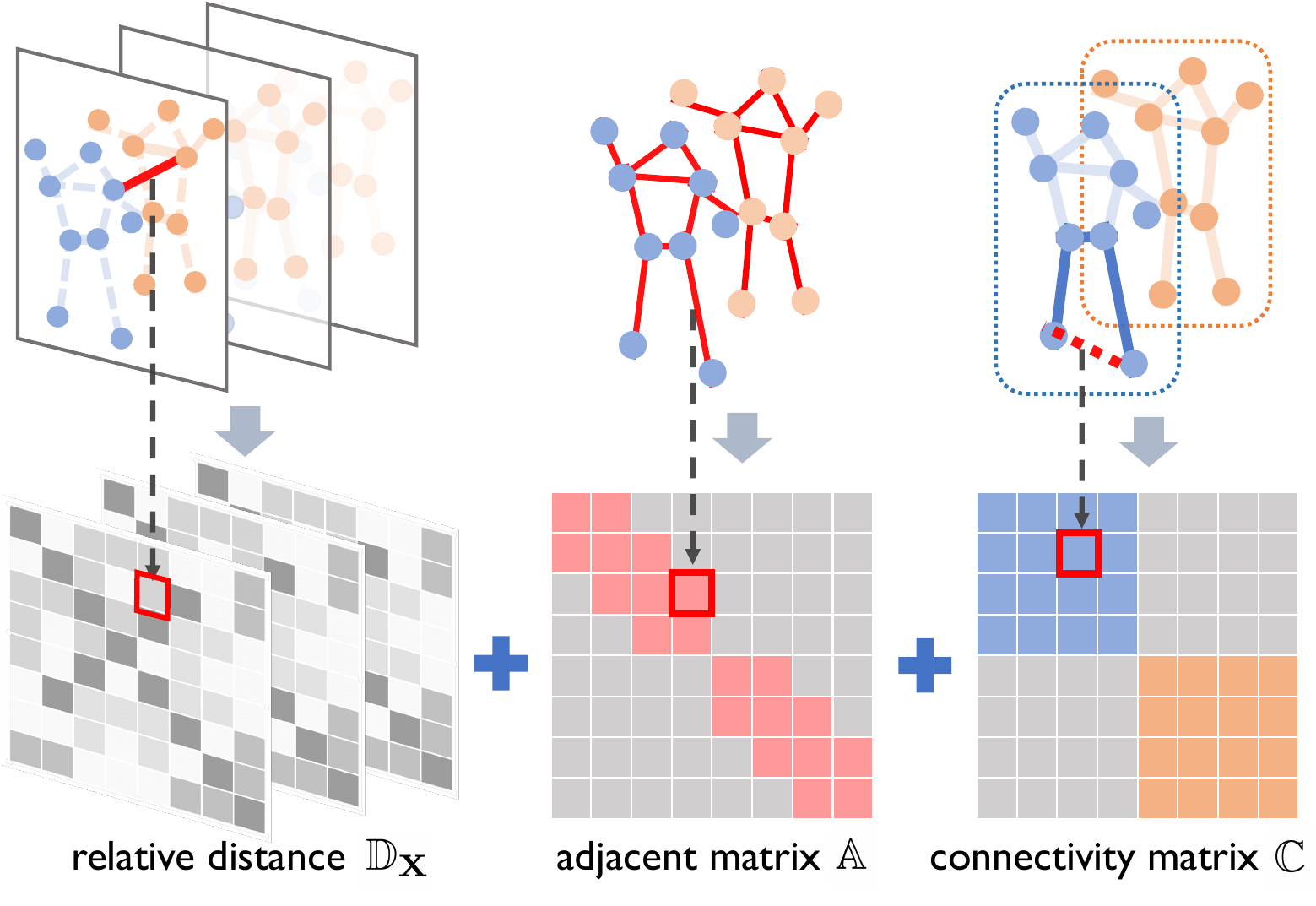}
\end{center}
   \caption{Three key components in the relation information including relative distance matrix sequence, bone adjacent matrix, and connectivity matrix. The upper is a schematic diagram of several relations, and the lower is the corresponding relation information.}
\vspace{-10pt}
\label{fig:realtion_information}
\end{figure}

\vspace{5pt}
\noindent \textbf{Relation Encoder.}
In addition to individually encoding joints, we also propose to encode the relation between joints from three perspectives: 
(i) to capture the relative distance between joints, we compute the distance information $\mathbb{D}_{\mathbf{X}}\in\mathbb{R}^{NJ\times NJ \times T_h}$ that consists of $T_h$ distance-aware matrices for each frame of the input sequence, where $(\mathbb{D}_{\mathbf{X}})_{ij}^t=e^{- \lVert X_i^t- X_j^t \rVert_2}$ is calculated as the negative exponent of the distance between the $i$-th and the $j$-th joint at the $t$-th frame, $X_i$ and $X_j$ represents the corresponding joint's position;
(ii) to reflect the skeleton structure  where joints connected by bone show stronger associations, we compute the adjacent matrix 
 $\mathbb{A}\in\mathbb{R}^{NJ\times NJ \times 1}$ where $\mathbb{A}_{ij}=1$ represents there is a bone connection between joint $i$ and $j$, and $0$ otherwise; 
 (iii) to model the connectivity between joints since those belonging to the same individual tend to exhibit similar overall movement patterns, we introduce a connectivity matrix $\mathbb{C}\in \mathbb{R}^{NJ\times NJ \times 1}$, where $\mathbb{C}_{ij}=1$ if there is a path defined on bones between joint $i$ and $j$. We refer to the combination of the last two items $\mathbb{A}$ and $\mathbb{C}$ as physical constraints between joints.

Therefore, the initial relation information can be defined as a tensor, 
{\em i.e.}, $\mathbb{R}_{\mathbf{X}}=[\mathbb{D}_{\mathbf{X}}, \mathbb{A}, \mathbb{C}]\in \mathbb{R}^{NJ\times NJ \times (T_h+2)}$, 
incorporating both historical distances information and physical constraints between each pair of joints, 
as shown in Fig.~\ref{fig:realtion_information}. 
Till here, the relation tensor is encoded by a $1\times 1$ convolution encoder, 
that fuses the $T_h+2$ channels and projects it to the same dimension as $\mathbf{F}_J$. 
The output relation feature is denoted as $\mathbb{F}^0_R\in \mathbb{R}^{NJ\times NJ\times D}$.



\begin{figure}[t]
\begin{center}
\vspace{-10pt}
\includegraphics[width=1.0\linewidth]{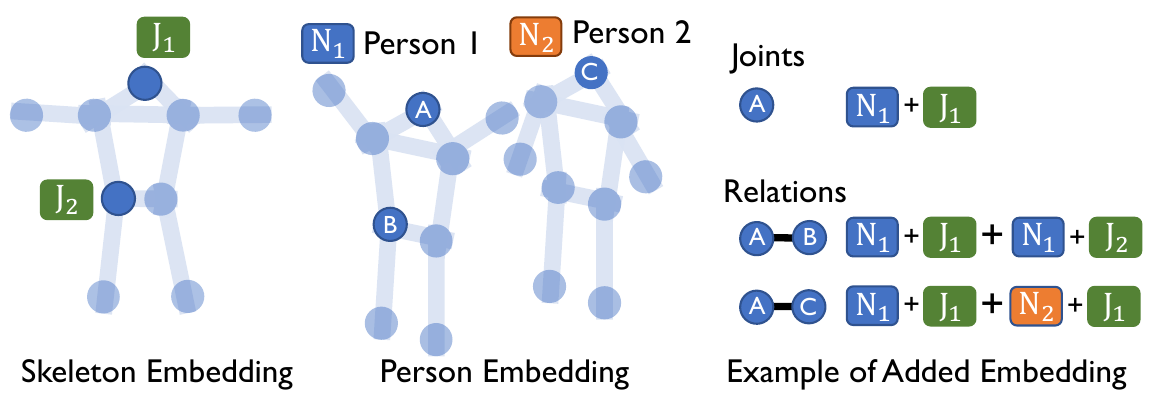}
\end{center}
   \caption{Example of joint embedding and relation embedding.}
\vspace{-12pt}
\label{fig:joint_embedding}
\end{figure}

\subsection{Fusion Module}
\label{sec:fusion_layer}
After separately encoding the joints and relations, 
their output features are further passed into a fusion module with positional information injected,
and followed by several novel Joint-Relation Fusion Layers, as detailed below.


\subsubsection{Positional Embedding}
We add the positional embedding to the output features from the joint and relation encoder respectively.
Specifically, for joint representation, we use $(N+J)$ learnable embeddings to indicate the person and joint identities respectively;
while for relation representation, the position embeddings are constructed by adding the positional embeddings of two related joints, as shown in Fig.~\ref{fig:joint_embedding}.
As a consequence, the resulting features from two-stream encoders become position-aware now.




\subsubsection{Joint-Relation Fusion Layer}
In this section, we describe two novel architectural designs that encourage the communication between joint and relation branches, namely, \textbf{relation-aware joint feature learning} and \textbf{joint-aware relation feature learning}. Specifically, the former enables to update the joints' features with the attention mechanism by querying information from the relation branch, and the latter further updates the relation features with message collection and local updates.




\vspace{5pt}
\noindent \textbf{Relation-Aware Joint Feature Learning.} 
To explicitly incorporate the relation information while updating joints' representation,
we design a novel relation-aware attention for joint feature learning, formulated as:
\begin{equation}
\label{eq:relation_aware_attn}
    \begin{aligned}
        &\mathbf{F}^{l+1}_J = f_{\rm{RA}}^l (\mathbf{F}^{l}_J, \mathbb{F}^{l}_R), \quad {\rm omit}\ l \ {\rm below \ for \ brevity},
        \\
        &\qquad={\rm MHA}(f_{\rm QKV-J}(\mathbf{F}_J), f_{\rm RF}(\mathbb{F}_R)), 
        \\
        &\qquad= {\rm MH} ({\rm softmax} (\dfrac{\mathbf{Q}_J (\mathbf{K}_J)^\top + f_{\rm RF}(\mathbb{F}_R)}{\sqrt{F_K}})\mathbf{V}_J).
    \end{aligned}
\end{equation}
Specifically, $\mathbf{F}_J^l \in\mathbb{R}^{NJ\times D}$ and $\mathbb{F}_R^l \in\mathbb{R}^{NJ\times NJ \times D}$ are the input joint and relation feature with added the positional embedding at the $l$-th layer. 
The relation function $f_{\rm RF}(\mathbb{F}_R) = \mathbb{F}_R\mathbf{W}_l +  \sum (\mathbb{F}_R\mathbf{W}_q^1 \odot \mathbb{F}_R\mathbf{W}_q^2)$ is composed with linear projection term and quadratic projection term where $\mathbf{W}_l\in\mathbb{R}^{D\times 1}, \mathbf{W}_q^1, \mathbf{W}_q^2\in\mathbb{R}^{D\times D'}$ are learnable parameters and we perform the sum operation among the last dimension.
${\rm MHA} (\cdot)$ refers to a variant of multi-head attention.
${\rm MH} (\cdot)$ denotes the operation of computing the weighted average of values at each head, and $F_K$ is the feature size of the key $\mathbf{K}_J$. 


In Eq.~\eqref{eq:relation_aware_attn}, we first generate the query/key/value for joint through the corresponding joint QKV encoder $f_{\rm QKV-J}(\cdot)$ and calculate the relation score through the relation function $f_{\rm RF}(\cdot)$ with the multi-head of $D_H$. We then fuse the relation and joint feature in the attention calculation, which adds the relation score to the initial attention score from the joint's query/key. We finally output the updated joint feature through the ${\rm softmax}(\cdot)$ and weighted sum operations. Note that here queries, keys, and values share the same feature size $D_K$. We adopt the quadratic term in the relation function for better representation ability.

We further perform local update that first employs a residual connection to the joint feature and then applies a feed-forward layer with a residual connection and a layer normalization to get the updated feature.

\vspace{5pt}
\noindent \textbf{Joint-Aware Relation Feature Learning.} 
With the fused and updated features through relation-aware attention, 
we elaborate the procedure for updating the relation feature by collecting all messages and performing the local update. 
Specifically, given $\mathbb{F}^{l}_R \in \mathbb{R}^{NJ\times NJ\times D}$ and $\mathbf{F}^{l+1}_J \in \mathbb{R}^{NJ\times D}$,
denoting the relation tensor and updated features for joints, respectively.
The operating procedure of the joint-aware relation feature learning can be formulated as:
\begin{subequations}
\label{eq:relation_update}
    \begin{align}
        \label{eq:ru_broadcast}
        &\mathbb{F}^{l+1}_J  = {\rm Broadcast} (\mathbf{F}^{l+1}_J) \in \mathbb{R}^{NJ\times NJ\times D},  
        \\
        \label{eq:ru_message_collect}
        & \mathbb{M}^{l+1} \!\!= \!\! [\mathbb{F}^{l+1}_J, (\mathbb{F}^{l+1}_J)^\top, \mathbb{F}^{l}_R, (\mathbb{F}^{l}_R)^\top] \! \in \! \mathbb{R}^{NJ\times NJ\times D_M},
        \\
        \label{eq:ru_local_update_1}
        & \mathbb{F}^{l+1}_R = \mathbb{F}^{l}_R + f_{\rm LU1} ({\rm Norm} (\mathbb{M}^{l+1}))\! \in \! \mathbb{R}^{NJ\times NJ\times D}, 
        \\
        \label{eq:ru_local_update_2}
        & \mathbb{F}^{l+1}_R = \mathbb{F}^{l+1}_R + f_{\rm LU2} ({\rm Norm} (\mathbb{F}^{l+1}_R))\! \in \! \mathbb{R}^{NJ\times NJ\times D},
    \end{align}
\end{subequations}
where $D_M = 4\times D$ is the hidden dimension of the collected message $\mathbb{M}^{l+1}$.
In detail, Step~\eqref{eq:ru_broadcast} broadcasts the joint feature to relations; Step~\eqref{eq:ru_message_collect} collects the message $\mathbb{M}^{l+1}$ for the current relation, which contains the joint features within the linked joints and bi-directional to-be update relation feature; Step~\eqref{eq:ru_local_update_1} performs the first round of local update with the normalization operation ${\rm Norm}(\cdot)$ and local update layer $f_{\rm LU}^1(\cdot)$; and Step~\eqref{eq:ru_local_update_2} performs the second round. 

\textbf{Note that}, in equations, (i) both update layers $f_{\rm LU1}(\cdot)$ and $f_{\rm LU2}(\cdot)$ are implemented through MLP, (ii) the transpose operation, {\em e.g.} $(\mathbb{F}^{l}_R)^\top$, is performed on the first two dimensions to collect the corresponding bi-directional features; and iii) we adopt the ${\rm LayerNorm}(\cdot)$ as the normalization operation.

\subsection{Decoder Module}
\label{sec:decoder}

Through the fusion module, the joint and relation features are well-fused in high dimensions. We now elaborate on the decoder module which projects the feature back to joint motion and distance between joints. 

\vspace{5pt}
\noindent \textbf{Joint Decoder.} In particular, we adopt the joint decoder $\mathcal{D}_J (\cdot)$ to decode the fused joint feature $\mathbf{F}^{L}_J \in \mathbb{R}^{NJ \times D}$ via a 3-layer MLP. 
It outputs the reconstructed joint movement $\widehat{\mathbf{X}}_{\text{NJ}} \in \mathbb{R}^{NJ \times (T_h\times 3)}$ over $T_h$ frames and predicted movement $\widehat{\mathbf{Y}}_{\text{NJ}} \in \mathbb{R}^{NJ \times (T_f\times 3)}$ over $T_f$ future frames at once. 

\vspace{5pt}
\noindent \textbf{Relation Decoder.} In order to better model the interaction from the perspective of distance changes between joints, we employ a $1\times 1$ convolution decoder $\mathcal{D}_R (\cdot)$ to decode the fused relation feature $\mathbb{F}^{L}_R \in \mathbb{R}^{NJ \times NJ \times D}$. In specific, the decoder fuses the $D$ feature channels and projects it to $T_h+T_f$ dimension where the first $T_h$ dimension is the reconstructed distance $\widehat{\mathbb{D}}_{\mathbf{X}}\in\mathbb{R}^{NJ \times NJ\times T_h}$ and the last $T_f$ represents the predicted distance $\widehat{\mathbb{D}}_{\mathbf{Y}}\in\mathbb{R}^{NJ \times NJ\times T_f}$.





\subsection{Training Objective}  \label{sec:training_objective}

To train the proposed Joint-Relation Model, 
we here adopt three types of supervision terms, 
including i) joint supervision on reconstructed and predicted joint motion, 
ii) relation supervision on reconstructed and predicted distance sequence, 
and iii) Transformer deep supervision on the input of each fusion layer, as detailed below.

\vspace{4pt}
\noindent \textbf{Joint Supervision.} 
Let $\widehat{\mathbf{X}}_{\text{NJ}}$ and $\widehat{\mathbf{Y}}_{\text{NJ}}$ be the reconstructed and predicted joint information and $\mathbf{X}_{\text{NJ}}$/$\mathbf{Y}_{\text{NJ}}$ be the corresponding ground-truth, we supervise them through
\begin{equation}
  \mathcal{L}_J (\widehat{\mathbf{X}}_{\text{NJ}}, \widehat{\mathbf{Y}}_{\text{NJ}}) = \lVert  \mathbf{X}_{\text{NJ}}-\widehat{\mathbf{X}}_{\text{NJ}}\rVert_2  + \lambda_J \lVert  \mathbf{Y}_{\text{NJ}}-\widehat{\mathbf{Y}}_{\text{NJ}}\rVert_2\nonumber,
\end{equation}
where $\lambda_J\in\mathbb{R}$ is a weight hyperparameter to balance two terms and $\lVert \cdot \rVert_2$ takes L2-norm on the 3D coordinates and average other dimensions.

\vspace{4pt}
\noindent \textbf{Relation Supervision.}  Let $\widehat{\mathbb{D}}_\mathbf{X}$ and $\widehat{\mathbb{D}}_\mathbf{Y}$ be the reconstructed and predicted distance and $\mathbb{D}_\mathbf{X}$/$\mathbb{D}_\mathbf{Y}$ be the corresponding ground-truth, we formulate the supervision as $$ \mathcal{L}_R (\widehat{\mathbb{D}}_\mathbf{X}, \widehat{\mathbb{D}}_\mathbf{Y}) = \lVert  \mathbb{D}_\mathbf{X}-\widehat{\mathbb{D}}_\mathbf{X}\rVert_1  + \lambda_R\lVert  \mathbb{D}_\mathbf{Y}-\widehat{\mathbb{D}}_\mathbf{Y}\rVert_1 ,$$
where $\lambda_R\in\mathbb{R}$ is the weight hyperparameter and $\lVert \cdot \rVert_1$ represents the average L1-norm between the ground truth distance and the predicted value.

\vspace{4pt}
\noindent \textbf{Deep Supervision.} To avoid overfitting caused by deep Transformer fusion layers, 
we use the same joint/relation decoder on the input of each fusion layer and calculate the corresponding losses:
\begin{equation}
  \mathcal{L}_{\rm DS}  = \sum_{l=0}^{L-1} \Big ( \mathcal{L}_J\big (\mathcal{D}_J (\mathbf{F}_J^l)\big ) + \mathcal{L}_R\big (\mathcal{D}_R (\mathbb{F}_R^l)\big ) \Big ).
\end{equation}

\vspace{4pt}
\noindent \textbf{Total Loss.} 
$\mathcal{L} = \mathcal{L}_J (\widehat{\mathbf{X}}_{\text{NJ}}, \widehat{\mathbf{Y}}_{\text{NJ}}) + \mathcal{L}_R (\widehat{\mathbb{R}}_\mathbf{X}, \widehat{\mathbb{R}}_\mathbf{Y}) + \mathcal{L}_{\rm DS}$.
\section{Experimental Setup}
\subsection{Datasets}

\begin{table*}[!t]
\small
\begin{center}
\caption{Experimental results in VIM on the 3DPW-SoMoF~(left) and 3DPW-SoMoF/RC~(right) test sets. The best results are highlighted in bold. Our method outperforms most previous state-of-the-art methods on 3DPW-SoMoF and achieves the best performance on the 3DPW-SoMoF/RC test sets.}
\vspace{3pt}
\label{tab:main_somof}
\begin{tabular}{l|cccccc|cccccc}
\toprule
\toprule
\multirow{2}{*}{Methods} & \multicolumn{6}{c}{3DPW-SoMoF}                & \multicolumn{6}{|c}{3DPW-SoMoF/RC}           \\
\cmidrule{2-13}
                  & AVG & 100 & 240 & 500 & 640 & 900 & AVG & 100 & 240 & 500 & 640 & 900 \\
\midrule 
Zero Velocity     & 86.7 & 29.4 & 53.6 & 94.5 & 112.7 & 143.1 & 67.2 & 23.9 & 42.9 & 73.1 & 86.2 & 109.7\\
LTD~\cite{wei2019motion} {\scriptsize{$^{\rm \textcolor{blue}{'2019}}$}} & 76.7 & 22.0 & 41.1 & 81.0 & 100.2 & 139.7 & 62.2 & 20.2 & 37.2 & 68.6 & 81.3 & 103.5\\
TRiPOD~\cite{adeli2021tripod}{\scriptsize{$^{\rm \textcolor{blue}{'2021}}$}}            & 84.2 & 31.0 & 50.8 & 84.7 & 104.1 & 150.4 & - & - & - & - & - & - \\
DViTA~\cite{parsaeifard2021decoupled}{\scriptsize{$^{\rm \textcolor{blue}{'2021}}$}}           & 65.7 & 19.5 & 36.9 & 68.3 & 85.5 & 118.2 &  57.4 & 15.0 & 32.3 & 62.7 & 76.3 & 100.6 \\
MRT~\cite{wang2021multi}{\scriptsize{$^{\rm \textcolor{blue}{'2021}}$}}  & 59.2 & 21.8 & 39.1 & 65.1 & 75.9 & 94.1 & 52.3 & 20.8 & 36.4 & 58.2 & 66.6 & 79.4 \\
FutureMotion~\cite{wang2021simple}{\scriptsize{$^{\rm \textcolor{blue}{'2021}}$}}      & 49.4 & 9.5 & 22.9 & 50.9 & 66.2 & 97.4 & - & - & - & - & - & -\\
SoMoFormer~\cite{vendrow2022somoformer}{\scriptsize{$^{\rm \textcolor{blue}{'2022}}$}}        & \textbf{46.3} & \textbf{9.1} & \textbf{21.3} & \textbf{47.5} & \textbf{61.6} & \textbf{91.9} & 43.8 & 11.0 & 24.8 & 49.1 & 59.2 & 74.8 \\ 
\midrule
\textbf{Ours}   & 47.2        & 9.5 & 22.1 &  48.7 & 62.8 & 92.8 & \textbf{39.5} & \textbf{9.5} & \textbf{21.7} & \textbf{44.1} & \textbf{53.4} & \textbf{68.8} \\
\bottomrule
\bottomrule
\end{tabular}
\vspace{-15pt}
\end{center}
\end{table*}

We evaluate our method on three multi-person motion datasets, including 3DPW~\cite{vonMarcard2018}, CMU-Mocap~\cite{cmu-mocap}, and MuPoTS-3D~\cite{singleshotmultiperson2018}.

\vspace{4pt}
\noindent \textbf{3DPW.} 3D Poses in the Wild Dataset~(3DPW) is a large-scale 3D motion dataset collected by moving mobile phone cameras with pose estimation and optimization~\cite{vonMarcard2018}. In this paper, we use the SoMoF benchmark~\cite{adeli2020socially,adeli2021tripod} splits~(3DPW-SoMoF), {\em i.e.}, the sequences that contain two persons, 
and predict future 900ms~(14 frames) motion using the historical 1030ms~(16 frames) motion.

\vspace{2pt}
\noindent \textbf{3DPW-SoMoF/RC.}
We find the camera movement in 3DPW dataset causes a serious unnatural drift on persons, 
affecting the modeling of multi-person interaction. 
Here, we subtract the estimated camera velocity for better interaction modeling and generate a new dataset, 
termed as 3DPW-SoMoF/RC. See more details in supplementary materials.

\vspace{2pt}
\noindent \textbf{CMU-Mocap.} The Carnegie Mellon University Motion Capture Database~(CMU-Mocap) contains a large number of single-person scenes and only limited set of scenes with two persons~\cite{cmu-mocap}. 
We use the training set and test set given in the paper~\cite{wang2021multi}, 
where each scene contains 3 persons, 
and are obtained by sampling from single-person scenes with multi-person scenes and mixing them together. 
We aim to predict future 3000ms~(45 frames) motion using the historical 1000ms~(15 frames) motion.

\vspace{2pt}
\noindent \textbf{MuPoTS-3D.}
 Multiperson Pose Test Set in 3D~(MuPoTS-3D) consists of over 8000 frames collected from 20 sequences with 8 subjects~\cite{singleshotmultiperson2018}. Following previous works~\cite{wang2021multi, vendrow2022somoformer}, 
 we evaluate our model's performance with the same segment length as CMU-Mocap on the test set.

\subsection{Metrics}

\noindent \textbf{VIM.} Visibility-Ignored Metric (VIM) is proposed in SoMoF benchmark~\cite{adeli2020socially, adeli2021tripod} to measure the displacement on the joint vector with the dimension of $J\times 3$. To be specific, the VIM for $t$-th frame is calculated as 
$$
\text{VIM}@ t= \frac{1}{N}\sum_{n=1}^N\sqrt{\sum_{j=1}^J( Y_{nj}^t-\widehat{Y}_{nj}^t)^2 }.
$$

\vspace{2pt}
\noindent \textbf{MPJPE.} Mean Per Joint Position Error~(MPJPE) is another commonly used metric in the fields of pose estimation and motion prediction, which calculates the average Euclidean distance between the predicted value and the ground truth of all joints, that is,
$$
\text{MPJPE} = \frac{1}{T_f}\frac{1}{N}\frac{1}{J}\sum_{t=1}^{T_f}\sum_{n=1}^N\sum_{j=1}^J\Vert Y_{nj}^t-\widehat{Y}_{nj}^t\Vert_2.
$$
We use this metric on CMU-Mocap and MuPoTS-3D.

\subsection{Implementation Details}

For 3DPW-SoMoF and 3DPW-SoMoF/RC datasets, 
we first pre-train the model on the AMASS~\cite{AMASS:ICCV:2019} dataset following previous works~\cite{wang2021simple, vendrow2022somoformer}, which provides massive motion sequences. 
We use the {\em CMU} subset as the training set and the {\em BioMotionLab}\_{\em NTroje} for test. 
We randomly sample single-person sequences from the dataset and mix them to get a synthetic set. 
While finetuning the model, we extend the 3DPW-SoMoF and 3DPW-SoMoF/RC datasets by sampling with overlap every two frames from the original 3DPW dataset. We perform three kinds of data augmentation including i) random rotation: the entire scene is rotated by a random angle within $[0, 2\pi]$ along the vertical axis; ii) person permutation: the person order in a scene is randomly permuted; and iii) sequence reverse: the entire sequence is temporally reversed and the last $T_h$ frames of the original sequence are taken as input. Here all the input sequences are normalized by subtracting the mean joint position of the first person in the first frame.

\begin{figure*}[ht]
\begin{center}
   \includegraphics[width=1.0\linewidth]{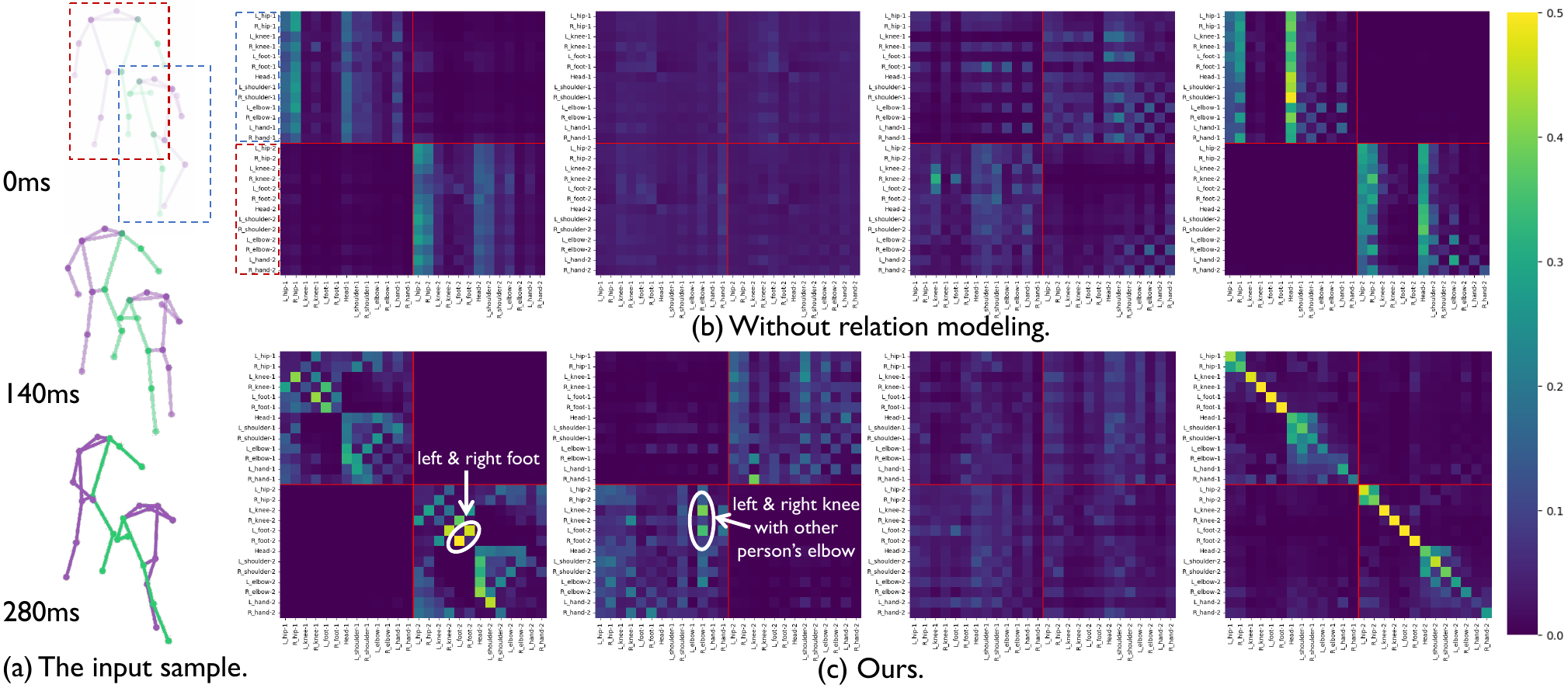}
\vspace{-20pt}
\end{center}
   \caption{Attention visualization in the first attention layer with different heads.~(a) shows the input sequence;~(b) and~(c) shows the plain/relation-aware attention matrices in the first layer. With the help of relation information, proposed relation-aware attention generates more reasonable attention allocations:~(from left to right) intra-body attention, inter-body attention, global attention, and self-attention.}
\label{fig:attn}
\end{figure*}

\vspace{4pt}
\noindent \textbf{Training Details.}
Our model has $L=4$ joint-relation fusion layers with $D_H=8$ attention heads, the feature dimension $D$ is set to 128. 
On 3DPW-SoMoF and 3DPW-SoMoF/RC datasets, we first pre-train for 100 epochs with an initial learning rate of $1\times 10^{-3}$ and decay by 0.8 every 10 epochs. 
When fine-tuning, our learning rate is set to $1\times 10^{-4}$ with a 0.8 decay every 10 epochs. 
The batch size is set to 128 for both pre-train and finetune. We use $\lambda_J=\lambda_R=10$. 
The whole network is implemented using Pytorch with AdamW optimizer.

\begin{table}[!t]
\small
\begin{center}
\caption{Experimental results in MPJPE on CMU-Mocap~(left) and
MuPoTS-3D~(right) test sets. The best results are highlighted in bold. Our method outperforms most state-of-the-art methods.}
\label{tab:cmu&mupost_result}
\setlength{\tabcolsep}{4pt}
\begin{tabular}{l|ccc|ccc}
\toprule
\toprule
\multirow{2}{*}{Methods} & \multicolumn{3}{c}{CMU-Mocap} & \multicolumn{3}{|c}{MuPoTS-3D} \\ 
\cmidrule{2-7}
                        & 1s       & 2s       & 3s      & 1s       & 2s       & 3s      \\ 
\midrule   
LTD~\cite{wei2019motion}{\scriptsize{$^{\rm \textcolor{blue}{'2019}}$}}                    & 13.7 & 21.9 & 32.6 & 11.9 & 18.1   & 23.4 \\
HRI~\cite{wei2020his}{\scriptsize{$^{\rm \textcolor{blue}{'2020}}$}}                   & 14.9 & 26.0 & 30.7 &  9.4  & 16.8 &  22.9 \\

MRT~\cite{wang2021multi}{\scriptsize{$^{\rm \textcolor{blue}{'2021}}$}}                     & 9.6 &  15.7 &21.8   & \textbf{8.9} & 15.9   & 22.2\\ 
SoMoFormer~\cite{vendrow2022somoformer}{\scriptsize{$^{\rm \textcolor{blue}{'2022}}$}}              &  10.7 & 17.2 & 22.5 & 10.1 & 17.7 & 24.2\\ 
\midrule 
\textbf{Ours}                    & \textbf{8.3} & \textbf{13.9} & \textbf{18.5} &  \textbf{8.9} & \textbf{15.5} & \textbf{21.3}\\
\bottomrule
\bottomrule
\end{tabular}
\vspace{-20pt}

\end{center}
\end{table}

\subsection{Quantitative results}


\vspace{4pt}
\noindent \textbf{Results on 3DPW-SoMoF and 3DPW-SoMoF/RC.}
We report the experimental results on the 3DPW-SoMoF and 3DPW-SoMoF/RC datasets in Tab.~\ref{tab:main_somof}. 
For a fair comparison, we use the same VIM criterion at multiple future frames.
On standard SoMoF benchmark~(left section of Tab.~\ref{tab:main_somof}),
our method still outperforms most all previous methods on the 3DPW-SoMoF dataset, despite of serious drift problems; on SoMof/RC benchmark with drifting removed~(right section of Tab.~\ref{tab:main_somof}), 
our method reduces the VIM at AVG/900ms from 43.8/74.8 to 39.5/68.8, compared to the current state-of-the-art method  SoMoFormer~\cite{vendrow2022somoformer}, achieving 9.8\%/13.4\% improvement, reflecting the effectiveness of the proposed method.




\vspace{4pt}
\noindent \textbf{Results on CMU-Mocap and MuPots-3D.}
We also compare the result on CMU-Mocap and MuPots-3D dataset between our method and MRT~\cite{wang2021multi}, SoMoFormer~\cite{vendrow2022somoformer} as well as two recent single-person motion prediction methods including HRI~\cite{wei2020his} and LTD~\cite{wei2019motion}. Both methods are trained on the synthesized dataset given by MRT and tested on the corresponding CMU-Mocap test set and MuPots-3D test set. We report the MPJPE result  on predicting 1, 2, and 3s motion as the MRT does in Tab.~\ref{tab:cmu&mupost_result}. We obverse a significant  performance improvement on CMU-Mocap and our method also achieves most of state-of-the-art results on MuPots-3D. 
This demonstrates the strong ability of our model to predict multi-person motion and strong generalization.

\subsection{Qualitative Results}

\vspace{4pt}
\noindent \textbf{Visualization of attention weights.}
To verify the effectiveness of the proposed relation-aware attention, we visualize the learned attention matrices in the first layer by the relation-aware attention and the plain attention over joints; see Fig.~\ref{fig:attn}. 
We see that our model has learned to explicitly generate diverse attention matrices including inter-person attention, 
intra-person attention, global attention, and self-attention for different heads.

\begin{figure}[t]
\begin{center}
   \includegraphics[width=1.0\linewidth]{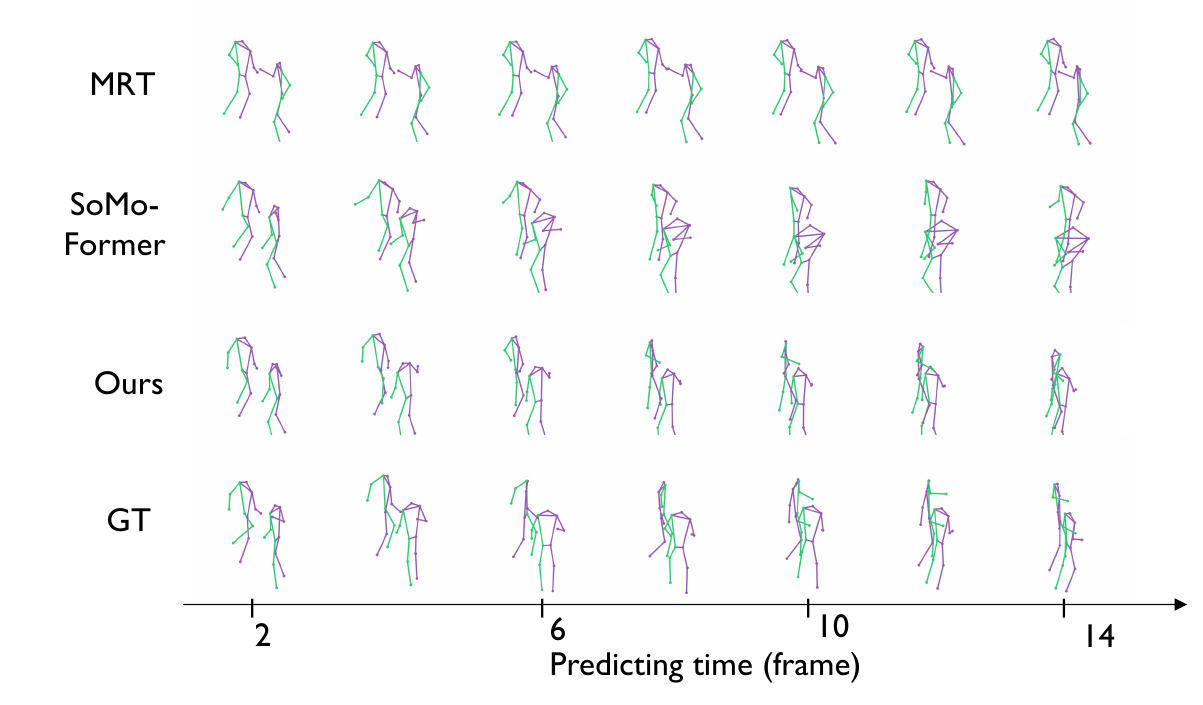}
\end{center}
   \caption{Visualization comparison on 3DPW-SoMoF/RC dataset. We compare the prediction by our method and two previous methods. Our method generates a more precise motion prediction.}
\label{fig:prediction}
\vspace{-10pt}

\end{figure}

\vspace{4pt}
\noindent \textbf{Visualization of prediction result.}
We provide the qualitative comparison between our method with other recent methods including MRT~\cite{wang2021multi} and SoMoFormer~\cite{vendrow2022somoformer}. We visualize the predicted motion sequences from the 3DPW-SoMoF/RC test set, see Fig.~\ref{fig:prediction}. Compared with MRT's predictions which tend to be static, and SoMoFormer's predictions where the human skeletons appear unnaturally distorted, our method generates predictions that are not only more vivid and natural but also structurally correct. The visualization results validate the performance of our model.

\subsection{Ablation Study}

\vspace{4pt}
\noindent \textbf{Relation information and relation supervision.}
To verify: i) the effect of the introduced relation information; 
ii) the rationality and effect of distance information and physical constraints information included in our designed relational information; and iii) the role of additional relation supervision, 
in this section, we study the performance of the model on the 3DPW-SoMoF/RC test set with different relation information inputs with or without relation supervision. 
Tab.~\ref{tab:relation_ablation} represents the VIM result with different settings on 3DPW-SoMof/RC. 
We can conclude that: i) incorporating additional relation information, {\em e.g.}, distance information or physical constraints or both, can significantly improve model performance, proving the effectiveness of relation information in better modeling the joints interaction and movements; 
ii) both the distance relation information and physical constraints relation information contribute to the improvement of model performance, while the best performance is achieved when both are considered; 
iii) the relation supervision is also helpful since it can help model capture the interaction information hidden in distance changes.

\begin{table}[!t]
\small
\begin{center}
\setlength{\tabcolsep}{3pt}
\caption{Ablation of relation information and relation supervision in VIM on 3DPW-SoMoF/RC test sets. Each notation is defined as {\em Dist.}: the relation information only include distance information $\mathbb{D}_{\mathbf{X}}$, {\em Phys.}: physical constraints include the adjacent matrix $\mathbb{A}$ and 
 the connectivity matrix $\mathbb{C}$. With both kinds of relation information and relation supervision,  the model performs best.}
\label{tab:relation_ablation}
\vspace{5pt}
\begin{tabular}{ccc|cccccc}
\toprule
\multicolumn{2}{c|}{Relation}                            & \multicolumn{1}{c|}{\multirow{2}{*}{\begin{tabular}[c]{@{}c@{}}Relation \\Supervision\end{tabular}}} & \multicolumn{6}{c}{3DPW-SoMoF/RC}                                                                                                                         \\ \cmidrule{1-2} \cmidrule{4-9} 
\multicolumn{1}{c}{{\em Dist.}} & \multicolumn{1}{c|}{{\em Phys.}} & \multicolumn{1}{c}{}                                                                                & \multicolumn{1}{|c}{AVG} & 100 & 240 & 500 & 600 & 900 \\
\midrule 
 & & & 41.7 & 10.3 & 23.4 & 46.8 & 56.7 & 71.5 \\
\checkmark & &                          & 41.1 & 9.8 & 22.8 & 46.2 & 56.0 & 71.1\\
& \checkmark&                           & 40.8 & 9.9 & 22.6 & 45.3 & 55.4 & 70.7\\
\checkmark & &\checkmark                & 39.9 & 9.6 & 21.8 & \textbf{43.9} & 54.0 & 70.1\\
& \checkmark& \checkmark                & 40.1 & 9.8 & 22.1 & 44.6 & 54.1 & 69.7\\
\checkmark & \checkmark&                & 40.3 & 9.8 & 22.4 & 45.0 & 54.6 & 69.8\\ \midrule
\checkmark & \checkmark  &\checkmark   & \textbf{39.5} & \textbf{9.5} & \textbf{21.7} & 44.1 & \textbf{53.4} & \textbf{68.8}\\
\bottomrule
\end{tabular}
\end{center}
\vspace{-10pt}
\end{table}

\vspace{4pt}
\noindent \textbf{Interaction modeling.}
To demonstrate the effectiveness of interaction modeling, we compare our model with other three settings: 
(i) single person model that uses the general self-attention to only capture the relationship between the intra-person joints; 
(ii) single person with relation model where the interaction between people is masked and we only provide and supervise the distance of joints inside the human body; 
(iii) simple multi-person model which replaces the joint-relation fusion layer in our model with the general self-attention layer. The results are shown in Tab.~\ref{tab:single_multi_ablation}. It is obvious that the interaction between persons contributes the improvement of motion prediction performance, while with the modeling and utilization of relational information, our model can better capture the interaction information to predict future motion.

\begin{table}[!t]
\small
\begin{center}
\caption{Ablation of interaction modeling in VIM on 3DPW-SoMoF/RC test sets. With the interaction modeling and relation information, the model achieves the best prediction performance.}
\label{tab:single_multi_ablation}
\vspace{5pt}
\setlength{\tabcolsep}{3pt}
\begin{tabular}{cc|cccccc}
\toprule
Joint & Relation & AVG & 100 & 240 & 500 & 600 & 900 \\
\midrule
single-person       &           & 46.2 & 10.1 & 23.7 & 49.8 & 62.0  & 85.4\\
single-person       & \checkmark&  45.1   &  9.6   & 22.6    &   48.0  &   60.1  &  85.1   \\
multi-person        &           & 41.7 & 10.3 & 23.4 & 46.8 & 56.7 & 71.5\\
multi-person        & \checkmark& \textbf{39.5} & \textbf{9.5} & \textbf{21.7} & \textbf{44.1} & \textbf{53.4} & \textbf{68.8} \\
\bottomrule
\end{tabular}
\vspace{-20pt}
\end{center}
\end{table}

\section{Conclusion}

This paper proposes the Joint-Relation Transformer, a two-stream Transformer-based architecture for multi-person motion prediction, which introduces the relation information and designs a novel relation-aware attention to inject the relation information into joint movement. Extensive experiments show that our method achieves state-of-the-art performance on three datasets and qualitative results show the effectiveness of the learned attention matrices. 


\vspace{4pt}
\textbf{Limitation and future work.} This work considers the deterministic multi-person motion prediction where the model only predicts once. A possible future work is to explore the stochastic multi-person motion prediction where the model is required to model the diverse future distribution. Also, we will further explore the multi-scale structure among multi-person to make a more precise prediction.  

\vspace{4pt}
\textbf{Acknowledgement.} This research is supported by NSFC under Grant 62171276 and the Science and Technology Commission of Shanghai Municipal under Grant 21511100900 and 22DZ2229005.


\clearpage
{\small
\bibliographystyle{ieee_fullname}
\bibliography{egbib}
}

\end{document}